\documentclass[10pt]{IEEEtran}

\usepackage[utf8]{inputenc}
\usepackage{graphicx}
\usepackage{amsmath}
\usepackage{amssymb}
\usepackage{subfigure}
\usepackage{amsthm}
\usepackage{url}
\usepackage[sort&compress,comma,numbers]{natbib}
\usepackage{etoolbox}

\title{Current-mode Memristor Crossbars for Neuromemristive Systems}
\author{Cory Merkel
\thanks{The author is with the Information Directorate, Air Force Research Laboratory, Rome,
NY, 13441-4505 USA e-mail: cory.merkel.1@us.af.mil}}

\makeatletter
\patchcmd{\@maketitle}
  {\addvspace{0.5\baselineskip}\egroup}
  {\addvspace{-1\baselineskip}\egroup}
  {}
  {}
\makeatother

\begin{document}


\maketitle

\begin{abstract}
Motivated by advantages of current-mode design, this brief contribution explores the implementation of weight matrices in neuromemristive systems via current-mode memristor crossbar circuits.  After deriving theoretical results for the range and distribution of weights in the current-mode design, it is shown that any weight matrix based on voltage-mode crossbars can be mapped to a current-mode crossbar if the voltage-mode weights are carefully bounded.  Then, a modified gradient descent rule is derived for the current-mode design that can be used to perform backpropagation training.  Behavioral simulations on the MNIST dataset indicate that both voltage and current-mode designs are able to achieve similar accuracy and have similar defect tolerance.  However, analysis of trained weight distributions reveals that current-mode and voltage-mode designs may use different feature representations.
\end{abstract}


\section{Introduction}

\IEEEPARstart{S}{teady} growth in memristor research and development is creating a frenzy of activity in the artificial neural network (ANN) community.  The excitement stems from the prospect of using memristors to build ANN application-specific integrated circuits--\textit{neuromemristive systems} (NMSs)--with unparallelled power and size efficiency.  Indeed, there have already been examples of NMS designs with orders-of-magnitude reduction in power consumption and area over software and non-memristor hardware implementations of ANNs \cite{Merkel2016}.  These improvements in efficiency are critical to the success of ANNs in size, weight, and power (SWaP)-constrained application domains such as unmanned aerial vehicles, personal health monitoring, spectrum management for handheld radios, and mobile devices.  However, the relative infancy of memristor research leaves many open questions and design challenges that need to be addressed for NMSs to reach their full potential.

One particularly important design aspect of NMSs is the circuitry used to drive the memristor devices.  A memristor is a 2-terminal non-volatile resistance switch, often built by sandwiching a defective oxide between two metal electrodes \cite{kuzum2013synaptic}.  The primary role of memristors in NMSs is to provide weighted connections, or synapses, between computational units, or neurons.  Modulation of a synapse's weight value is acheived by applying appropriately large, or superthreshold voltages to the memristors to change their conductance values.  In addition, combining memristors into high-density crossbar circuits provides an efficient way to represent an entire matrix of weights between layers of neurons.  Since memristors follow a state-dependent Ohm's law \cite{LChua2011}, a clear design choice is for each neuron's output to be represented by a voltage, so that the current passed from one neuron $i$ to another $j$ is given by $s_{i}=w_{i,j}x_{j}$, where $s$, $w$, and $x$ are unitless (normalized) versions of the current, conductance, and voltage, respectively.  This type of voltage-mode design has been investigated intensely in NMSs \cite{indiveri2013integration,burr2017neuromorphic}, but it suffers from a few drawbacks.  
First, each pre-synaptic neuron drives post-synaptic neurons through synapses, which can be modeled as parallel memristor conductance paths.  If the memristor conductances are large, then this places significant loading on the pre-synaptic neuron, limiting fanout and output swing.    Although this could be mitigated by adding buffers between the neuron and each synapse, it would lead to large area and power overheads.  Another drawback of voltage-mode neurons is that their output depends on the load that they are driving.  This means that the neuron's output is not only a function of its inputs, but also a function of the synapses it's driving, which is contradictory to the way ANNs operate.  Finally, long distance communication of continuous analog voltages presents a challenge since routing paths introduce signal diminishing voltage drops.

Previous work has identified some methods to address the challenges discussed above.  Synapse circuits that use multiple memristors have larger input impedance, reducing the load on voltage-mode neurons \cite{adhikari2015circuit,linn2010complementary}.  However, these circuits are $> \text{7}\times$ the size of a single memristor, and they aren't easily configured into high-density crossbar architectures.  Others have proposed 
spike-based \cite{wang2015energy} neurons which can be easily buffered using low-power digital circuits, but at the expense of increased neuron circuit complexity and latency.  Another approach is to use current-mode neurons to drive memristor-based synapses.  Current-mode circuits do not suffer from the same loading problems as the voltage-mode circuits discussed above.  In addition, current-mode circuits are generally able to operate at lower supply voltages and typically can achieve higher bandwidths, sometimes approaching the MOSFET intrinsic frequency $f_{T}$ \cite{Toumazou1990}.  Plus, currents are easily buffed using current mirrors, and several complex operations can be computed using simple current-mode circuits by taking advantage of translinear design principles \cite{Gilbert1975,chicca2014neuromorphic}.  Unfortunately, current-mode techniques have only been studied in non-crossbar-based NMSs \cite{merkel2014current}.  For the first time, this brief contribution explores the use of current-mode memristor crossbars for NMSs.  The rest of this paper proceeds as follows:  Section \ref{section:current_driven_memristor_crossbar} discusses voltage-mode and current-mode memristor crossbars, comparing the impact on weight range and distribution.  Sections \ref{section:dummy} and \ref{section:bipolar} describe methods for controlling the magnitude and sign of weights in a current-mode crossbar.  Section \ref{section:gradient_descent} proposes a modified training algorithm for current-mode crossbar NMSs, and Section \ref{section:mnist} shows behavioral simulation results for the proposed design on the MNIST dataset.  Section \ref{section:conclusions} concludes this work.

\section{Current-mode Memristor Crossbar}
\label{section:current_driven_memristor_crossbar}

\begin{figure}[!t]
\centering
\hspace{-5mm}
\subfigure[]{
\includegraphics[scale=0.55]{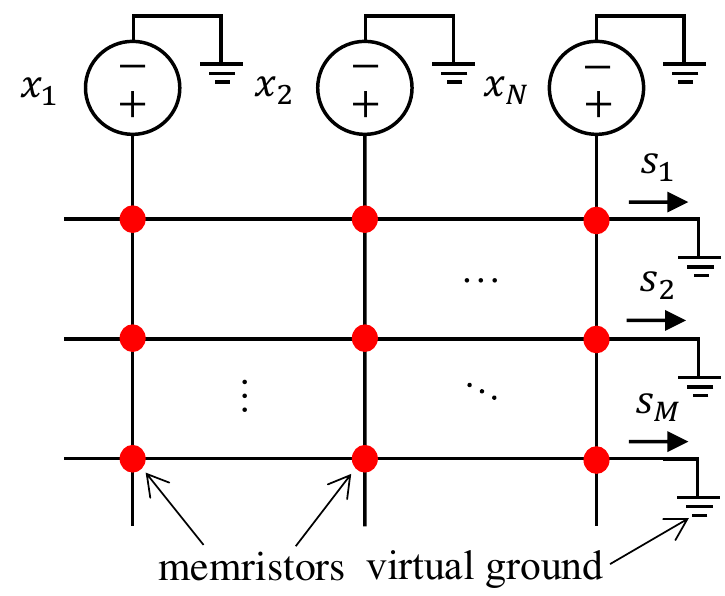}
\label{fig:voltage_xbar}
}
\hspace{-1mm}
\subfigure[]{
\raisebox{0.5mm}{
\includegraphics[scale=0.55]{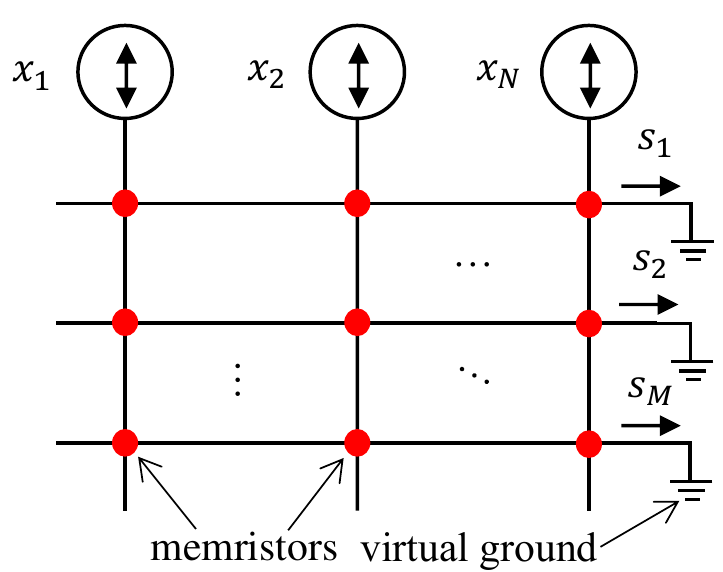}
\label{fig:current_xbar}
}}
\label{fig:xbars}
\caption{(a) Voltage-mode and (b) current-mode memristor crossbar configurations.}
\end{figure}

Matrix-vector products are ubiquitous in ANNs and can be implemented efficiently in NMSs using memristor crossbar circuits.  Consider the calculation
\begin{equation}
\mathbf{s}=\mathbf{W}\mathbf{x},
\label{eqn:s} 
\end{equation}
where $\mathbf{W}$ is an $M\times N$ weight matrix, and $\mathbf{x}$ is an $N$-dimensional column vector.  In addition, $x_{j}$ and $s_{i}$ are voltages and currents normalized to a real voltage or current denoted as $V_{max}$ or $I_{max}$.  The widely-adopted voltage-mode memristor crossbar configuration for implementing (\ref{eqn:s}) is shown in Figure \ref{fig:voltage_xbar}.  Each memristor, represented as a circle, has a conductance value $G$ that can be tuned within a particular range $G_{min}\le G\le G_{max}$.  In this work, $G_{min}=2.1\times 10^{-5}\mho$ and $G_{max}=1\times 10^{-3}\mho$, based on the memristors in \cite{Oblea2010}.  Note that, for both voltage-mode and current-mode memristor crossbars, it is desirable to have a small $G_{min}$ and a large $g\equiv G_{min}/G_{max}$ value in order to reduce sneak path currents and maximize distinguishability between memristor states.  Each crossbar row is connected to a virtual ground, which can be implemented with an opamp circuit.  This is a common technique used to reduce unintended current paths (sneak paths).  From Ohm's law, the current flowing through each memristor is equal to $G_{i,j}x_{i}$.  Furthermore, all of the currents that flow through memristors in a particular row will summate,
so the weight matrix represented by the voltage-mode crossbar circuit is just the matrix of memristor conductances, $w_{i,j}\equiv G_{i,j}/G_{max}$.

Now, consider the current-mode memristor crossbar circuit, shown in Figure \ref{fig:current_xbar}, where inputs to the circuit are currents that drive each column.  The current flowing through a memristor in a particular column is proportional to the memristor's conductance relative to the total memristor conductance in that column.  It is important to note that, in order to read the crossbar (i.e. perform the matrix-vector multiplication) without disturbing the memristor states, the input currents should be less than the memristor switching threshold $V_{th}$ times the sum of the conductances in each column:  $I_{max}<V_{th}\sum\limits_{i} G_{i,j}$. 
In the current-mode crossbar, the weight matrix is defined as:
\begin{equation}
w_{i,j}\equiv\frac{G_{i,j}}{\sum\limits_{k=1}^{M}G_{k,j}}.
\label{eqn:w_current_xbar}
\end{equation}
This redefinition of $w_{i,j}$ has a number of important implications on the range and distribution of weights that can be achieved in a current-mode crossbar.

\subsection{Weight Range}

In the voltage-mode case, the weight range is $g \le w_{i,j}\le 1$.  Compare this to the current-mode crossbar's weight range:  
\begin{equation}
\frac{1}{(M-1)g+1}\le w_{i,j}\le\frac{g}{M-1+g}.
\end{equation}
The weight range shrinks as the neuron fanout grows.  Note, however, that the dynamic range (ratio of maximum to minimum weight value) remains approximately constant over $M$.  Also, note that, for a given $M$, as the conductance ratio $g$ becomes large, the weight range approaches $0<=w_{i,j}<=1$.

\begin{figure}[!t]
\centering
\includegraphics[width=0.55\columnwidth]{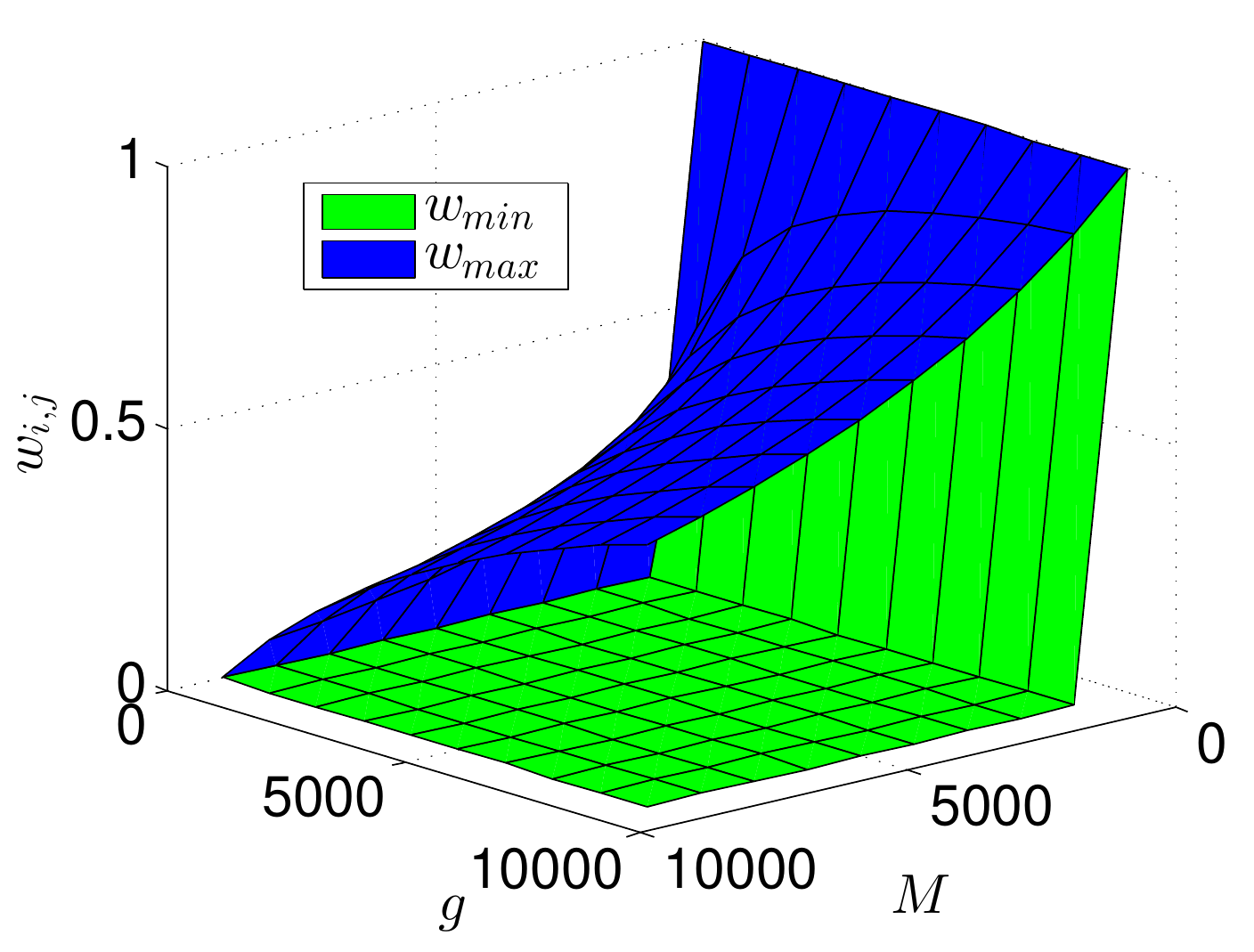}
\caption{Range of weights that can be implemented on the current-mode crossbar in Figure \ref{fig:current_xbar}.}
\label{fig:current_xbar_wrange}
\end{figure}

The value of the neuron fanout $M$ is application-dependent.  First, consider the current-mode crossbar in the output layer of a neural network used to classify handwritten digits.  In that case, $M=10$ (one output neuron per class).  In other situations, $M$ may be much larger.  For example, crossbars used as weight matrices in neural network hidden layers may have very large values of $M$.  In the human brain, $M$ is, on average, around 7000 to 10000.  Figure \ref{fig:current_xbar_wrange} shows the possible weight values for different values of $g$ and $M$.  For large neuron fanout, $g$ must be quite large to achieve a large maximum weight value.  However, if necessary, this limitation can be overcome by splitting the weight matrix across multiple crossbar circuits.

\subsection{Weight Distribution}
 
An interesting consequence of the current-mode crossbar configuration is that a constraint is placed on the weight distribution.  In the case of the voltage-mode design, any arbitrary weight matrix can be programmed on the voltage-mode crossbar as long as each weight falls within the range discussed above.  This isn't possible in the current-mode case, since the weights in a given crossbar column must sum to unity.  Let $\overline{\mathbf{W}}$ represent the target weight matrix to be mapped to the current-mode crossbar.  $\mathbf{W}$ is made as close as possible to $\overline{\mathbf{W}}$ by minimizing the element-wise differences:
\[
\begin{array}{ll}
\text{minimize}&\;\sum\limits_{i=1}^{M}\sum\limits_{j=1}^{N}\left(w_{i,j}-\overline{w}_{i,j}\right)^{2}\\[2ex]
\text{subject to}&\;\sum_{i}w_{i,j}=1, \forall j=1,2,\ldots,N.\\[1ex]
&\;w_{min}\le w_{i,j}\le w_{max}
\end{array}
\]
Using the method of Lagrange multipliers yields the solution:
\begin{equation}
w_{i,j}=\frac{1}{M}\left(1-\sum\limits_{k=1}^{M}\overline{w}_{k,j}\right)+\overline{w}_{i,j}.
\label{eqn:w_optimal}
\end{equation}
A geometrc interpretation is that we are finding $N$ points on the hyperplanes specified by $\sum_{i}w_{i,j}=1\forall j=1,1,\ldots,N$ which are closest to the $N$ points specified by the column vectors of $\mathbf{\overline{W}}$.  Now, to satisfy the inequality constraints, each point must lie in or on the hypercubes specified by $w_{min}\le w_{i,j}\le w_{max}$.  If the initial points lie outside of this hypercube, then they will be moved to the surface of the hypercube in the following manner.  First, the dimension of the initial point that is furthest outside the range of $[w_{min},w_{max}]$  is set to either $w_{min}$ or $w_{max}$.  Calculate $(1-\sum_{i}w_{i,j})/(M-1)$ and add this value to the remaining weights.  This process is repeated until the new point lies on the intersection of the hyperplane and hypercube.
Finally the conductances necessary to satisfy the weights are
\begin{equation}
G_{i,j}=\frac{w_{i,j}}{w_{k,j}}G_{k,j}
\end{equation}
This is an underdetermined system and, therefore, doesn't have a unique solution.  This makes sense intuitively since it is only ratios of conductances that determine weight values.  For example, a crossbar column where every conductance is equal to $G_{max}$ is equivalent to a crossbar column where all of the conductances are equal to $G_{min}$, or any other conductance value within the allowable range.

\begin{figure}[!t]
\centering
\includegraphics[scale=0.5]{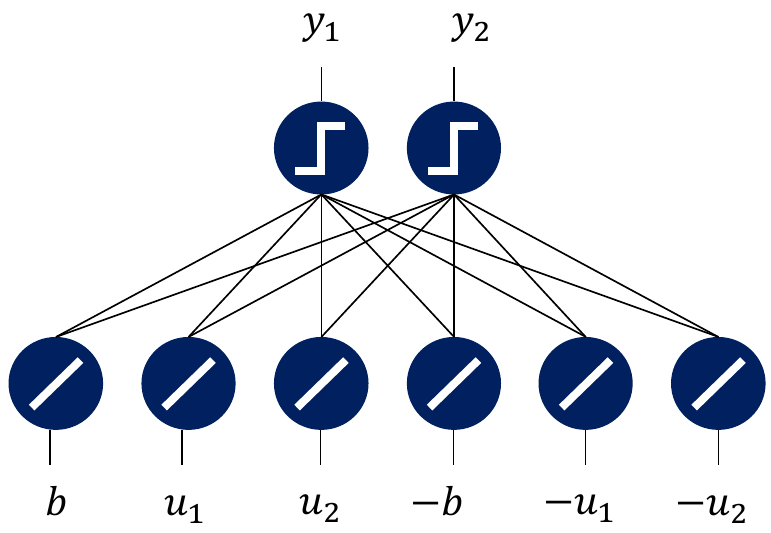}
\caption{Perceptron for computing AND and OR logic functions.}
\label{fig:logic_network}
\end{figure}

\begin{figure}[!t]
\centering
\subfigure[]{
\includegraphics[width=0.43\columnwidth]{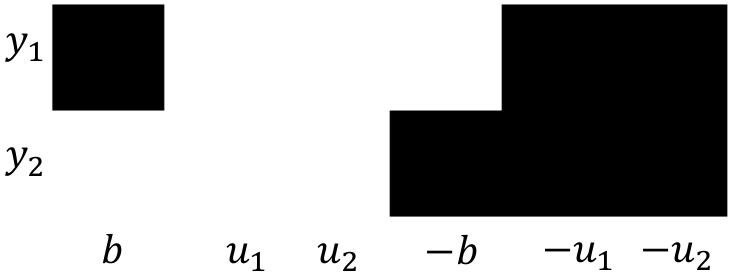}
\label{fig:logic_network_volt_weights}
}
\hspace{4mm}
\subfigure[]{
\includegraphics[width=0.43\columnwidth]{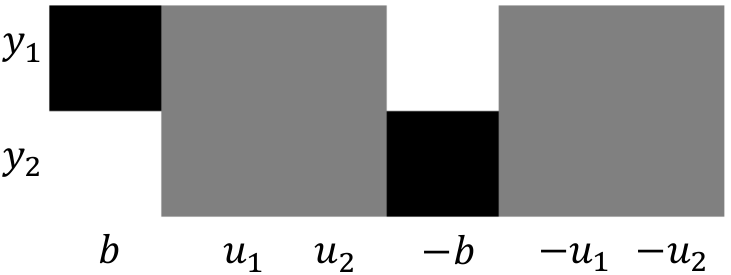}
\label{fig:logic_network_curr_weights}
}
\caption{Visualization of weight values for (a) a target weight matrix and (b) the optimized current-mode weight matrix when implementing AND and OR logic functions.}
\end{figure}

The above scheme was tested on a single-layer perceptron network trained to compute AND and OR logic functions (Figure \ref{fig:logic_network}).  All simulations are behavioral and performed in MATLAB.  Inputs were -1 and 1, and outputs were 0 and 1.  Note that true and negated inputs are required since the original weight definition in (\ref{eqn:w_current_xbar}) does not allow for negative weights.  However, with the negated values, each input has an effective weight of $w_{i,j}^{*}=w_{i,j}^{+}-w_{i,j}^{-}$, which will range from
\begin{equation}
w_{min}-w_{max}\le w_{i,j}^{*}\le w_{max}-w_{min}.
\end{equation}
A visual representation of the target weight matrix is shown in Figure \ref{fig:logic_network_volt_weights}.  The target matrix was found through the perceptron learning rule, with the weights bounded between 0 and 1.  Then, the weights were mapped to the current-mode crossbar, resulting in the weight matrix shown in Figure \ref{fig:logic_network_curr_weights}.  Lighter values correspond to larger weights and vice versa.  Many of the current-mode crossbar weight values are very different from the target matrix.  In this particular case, it occurs when the sum of the weights in the columns are small (i.e. all black) or large (i.e. all white).  If the sum of the target column weights is small, then they have to be scaled up to sum to unity.  Similarly, if the sum of the column weights is large, then they have to be scaled down.  This results in the 0.5 weight values (gray) in the crossbar.  The weight mismatch led to a 50\% classification error rate in the crossbar implementation, compared to perfect classification for the target.

\section{Controlling Weight Magnitudes by Adding a Dummy Crossbar Row}
\label{section:dummy}

\begin{figure}[!t]
\centering
\includegraphics[width=0.5\columnwidth]{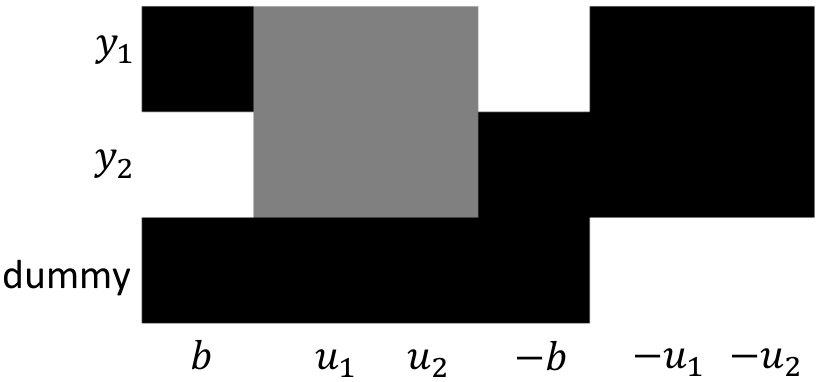}
\caption{Visualization of the current-mode crossbar weight matrix with an added dummy row.  The first two rows of the matrix represent the weights for implementing AND and OR logic functions.}
\label{fig:logic_network_curr_dummy_weights}
\end{figure}

A consequence of the current-mode crossbar is that the weights in a particular column can not be simultaneously controlled in terms of their relative and absolute magnitudes.  This leads to sub-optimal weight matrices as in the example above.  One remedy is to add a ``dummy" row to the crossbar.  All of the weights in a column of the current-mode crossbar weight matrix have to add to unity.  So, adding a ``dummy" weight to each column enables the control of the absolute size of the rest of the weights.  That is, if the dummy weight is large, then the rest of the weights will have to be small and vice versa.  The optimization problem is set up in a similar way as in the last section, resulting in
\begin{equation}
w_{i,j}=\overline{w}_{i,j}\forall i<M
\label{eqn:w_wbar}
\end{equation}
and
\begin{equation}
w_{M,j}=1-\sum\limits_{k=1}^{M}\overline{w}_{k,j}
\label{eqn:w_mplus1}
\end{equation}
This result is quite obvious.  The weights are being copied from the target matrix, and then the dummy weights are being set to whatever values are required for the columns to sum to unity.  The result for the AND and OR logic functions is shown in Figure \ref{fig:logic_network_curr_dummy_weights}.  Compare the top two rows to the unconstrained weight matrix in Figure \ref{fig:logic_network_volt_weights}.  The weights applied to $-u_{1}$ and $-u_{2}$ can now be matched by increasing the corresponding dummy weights.  However, the weights corresponding to $u_{1}$ and $u_{2}$ are still not able to match the target values, which are all large, resulting in 25\% error.  This would require the corresponding dummy weights to be negative, which isn't possible from the way the weights have been defined in (\ref{eqn:w_current_xbar}).

One possible workaround is to bound the target weight values so that (\ref{eqn:w_wbar}) and (\ref{eqn:w_mplus1}) result in weights that are within the range $[w_{min},w_{max}]$.  This results in
\begin{equation}
\mathrm{max}\left(\frac{1-w_{max}}{M-1},w_{min}\right)\le \overline{w}_{i,j}\le\mathrm{min}\left(\frac{1-w_{min}}{M-1},w_{max}\right).
\end{equation}
With these new constraints, the weights are gauranteed to match the target weights, leading to 100\% classification for the AND and OR logic problem.


\section{Bipolar Weights}
\label{section:bipolar}

One of the apparent drawbacks of using memristors to implement neural network weights is that their conductance is strictly positive (with the exception of negative differential resistance devices).  This makes it challenging to implement bipolar weights, which can be positive, negative, or zero.  One solution, which has been presented already, is to provide both the true and negated version of every crossbar input, leading to an effective bipolar input.  The drawback of this approach is that it requires 2$\times$ the crossbar size as for unipolar (e.g. strictly positive) weights.  Another solution is to subtract a fraction $\theta$ of the total pre-synaptic neuron output from each post-synaptic neuron, which can be done outside of the crossbar using current mirrors.  The total input to the pre-synaptic neuron becomes
\begin{equation}
s_{i}=\sum\limits_{j=1}^{N}\frac{G_{i,j}}{\sum\limits_{k=1}^{M}G_{k,j}}x_{j}-\theta\sum\limits_{j=1}^{N}x_{j}
\end{equation}
Therefore, the new effective weight value becomes
\begin{equation}
w_{i,j}^{*}=w_{i,j}-\theta
\end{equation}
The optimal value of $\theta$ is application-dependent, but for now it will be set so the target weight range is symmetric about 0, or
\begin{equation}
\theta=\frac{\overline{w}_{max}+\overline{w}_{min}}{2}.
\end{equation}

\section{Gradient Descent}
\label{section:gradient_descent}

\begin{figure}
\centering
\includegraphics[width=0.65\columnwidth]{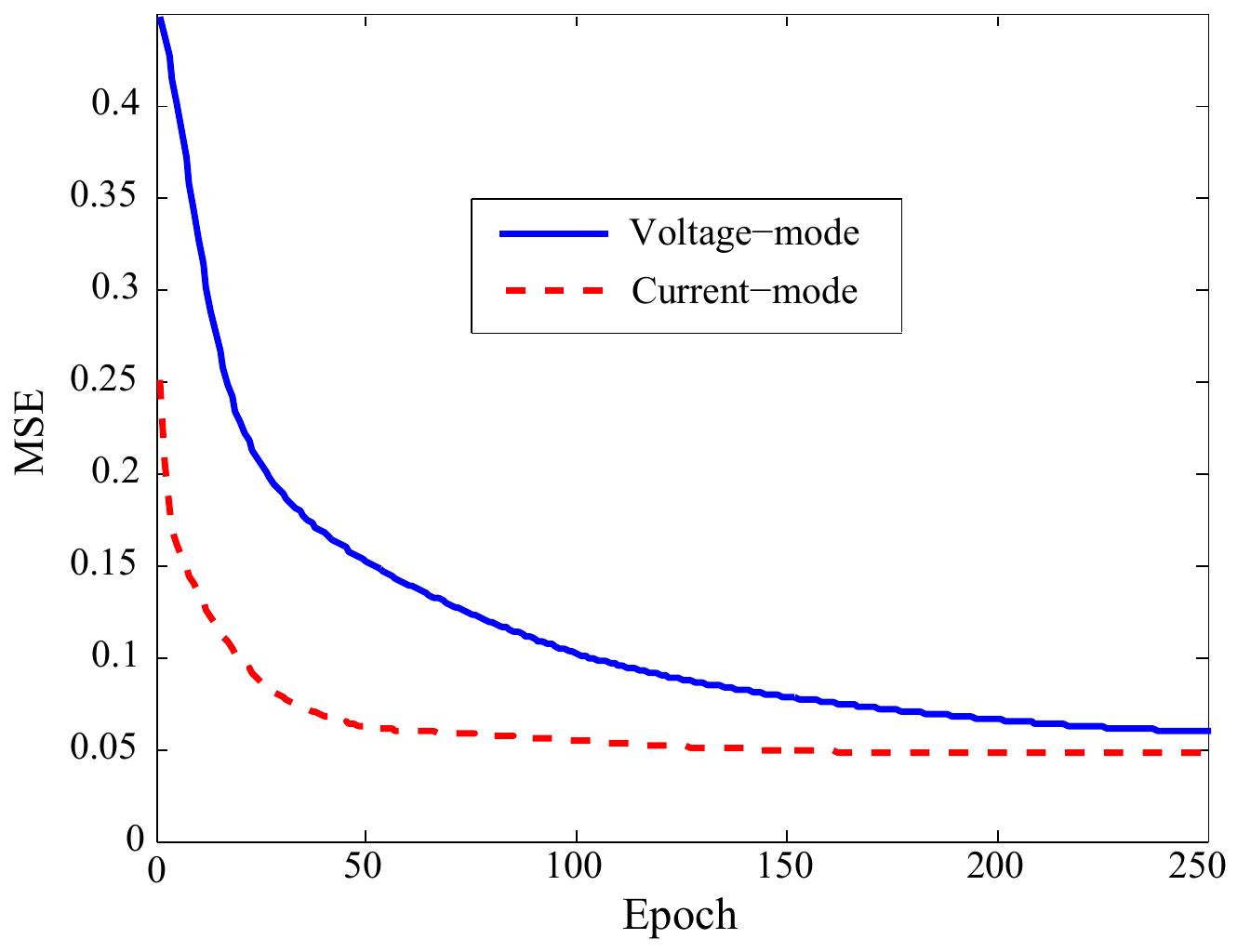}
\caption{MSE vs. epoch for gradient descent on the AND and OR logic problem.}
\label{fig:and_or_grad_descent}
\end{figure}

Let us now look at how to perform gradient descent (e.g. backpropagation) in a neural network that employs a current-mode memristor crossbar.  This amounts to minimizing a cost function $J$ with respect to the crossbar conductances:
\begin{equation}
\Delta G_{i,j}^{(l)}=-\alpha\frac{\partial J}{\partial G_{i,j}^{(l)}}=-\alpha\sum\limits_{k=1}^{M^{\prime(l)}}\frac{\partial J}{\partial x_{k}^{(l)}}\frac{\partial x_{k}^{(l)}}{\partial s_{k}^{(l)}}\frac{\partial s_{k}^{(l)}}{\partial w_{k,j}^{(l)}}\frac{\partial w_{k,j}^{(l)}}{\partial G_{i,j}^{(l)}},
\label{eqn:backprop}
\end{equation}
where $M^{\prime}$ is the number of non-dummy crossbar rows, which may be $M$ or $M-1$, and $l$ indexes the neural network layer.  Equivalently, (\ref{eqn:backprop}) can be written as
\begin{equation}
\Delta G_{i,j}^{(l)}=\alpha\sum\limits_{k=1}^{M^{\prime(l)}}\delta_{k}^{(l)}x_{j}^{(l-1)}\frac{\partial w_{k,j}^{(l)}}{\partial G_{i,j}^{(l)}}.
\label{eqn:dg}
\end{equation}
Expanding (\ref{eqn:dg}) yields a complicated, piecewise expression, but it is well-approximated by the simplified expression:
\begin{equation}
\Delta G_{i,j}^{(l)}\approx\left\{
\begin{array}{lr}
\alpha\delta_{i}^{(l)}x_{j}^{(l-1)} & : i\ne M\\[2em]
-\left<\Delta G_{1:M^{\prime},j}^{(l)}\right> & : i=M,
\end{array}
\right.
\label{eqn:current_update}
\end{equation}
\noindent
where $\left<\cdot\right>$ indicates arithmetic mean.  Note that for all of the non-dummy crossbar rows, this is just the standard backpropagation delta rule.  Implementing this gradient descent scheme on the AND and OR logic network yields the mean squared error (MSE) vs. training epoch curve shown in Figure \ref{fig:and_or_grad_descent}.  The corresponding voltage-mode crossbar-based network, trained with the standard gradient descent method, is shown for comparison.  Both gradient descent schemes show similar convergence.

\section{MNIST Experiments}
\label{section:mnist}

\begin{figure}[!t]
\centering
\includegraphics[width=0.63\columnwidth]{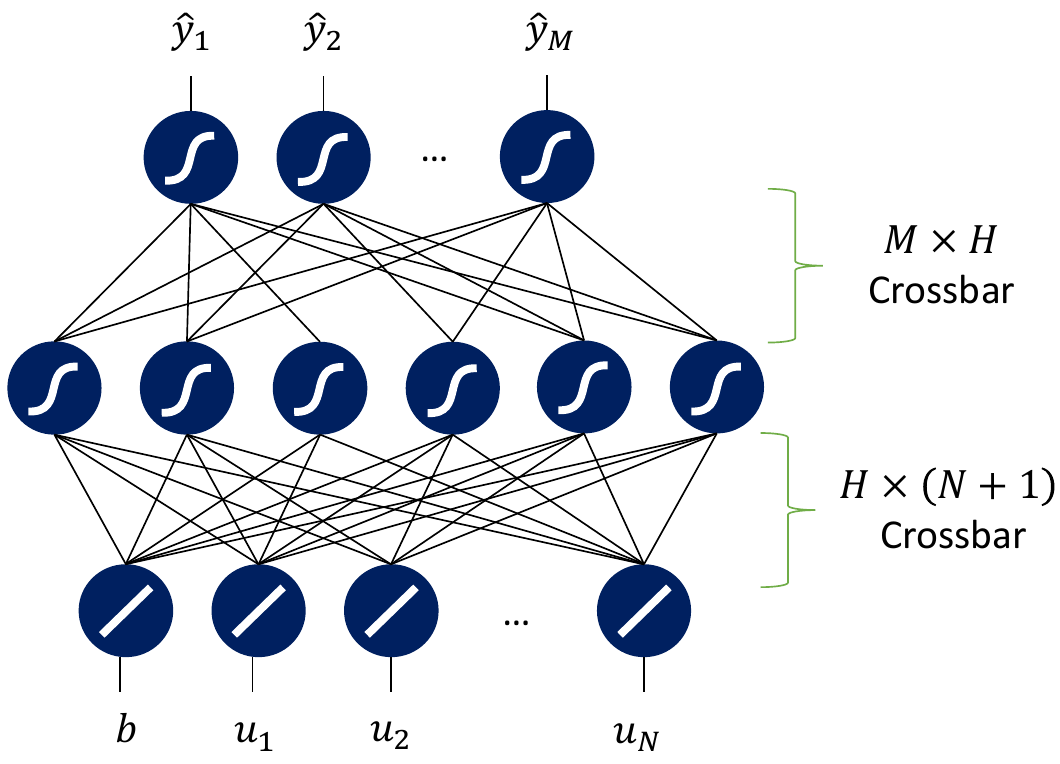}
\caption{MLP implemented with memristor crossbar-based weight matrices.}
\label{fig:mlp}
\end{figure}

\begin{figure}[!t] 
\centering
\includegraphics[width=0.65\columnwidth]{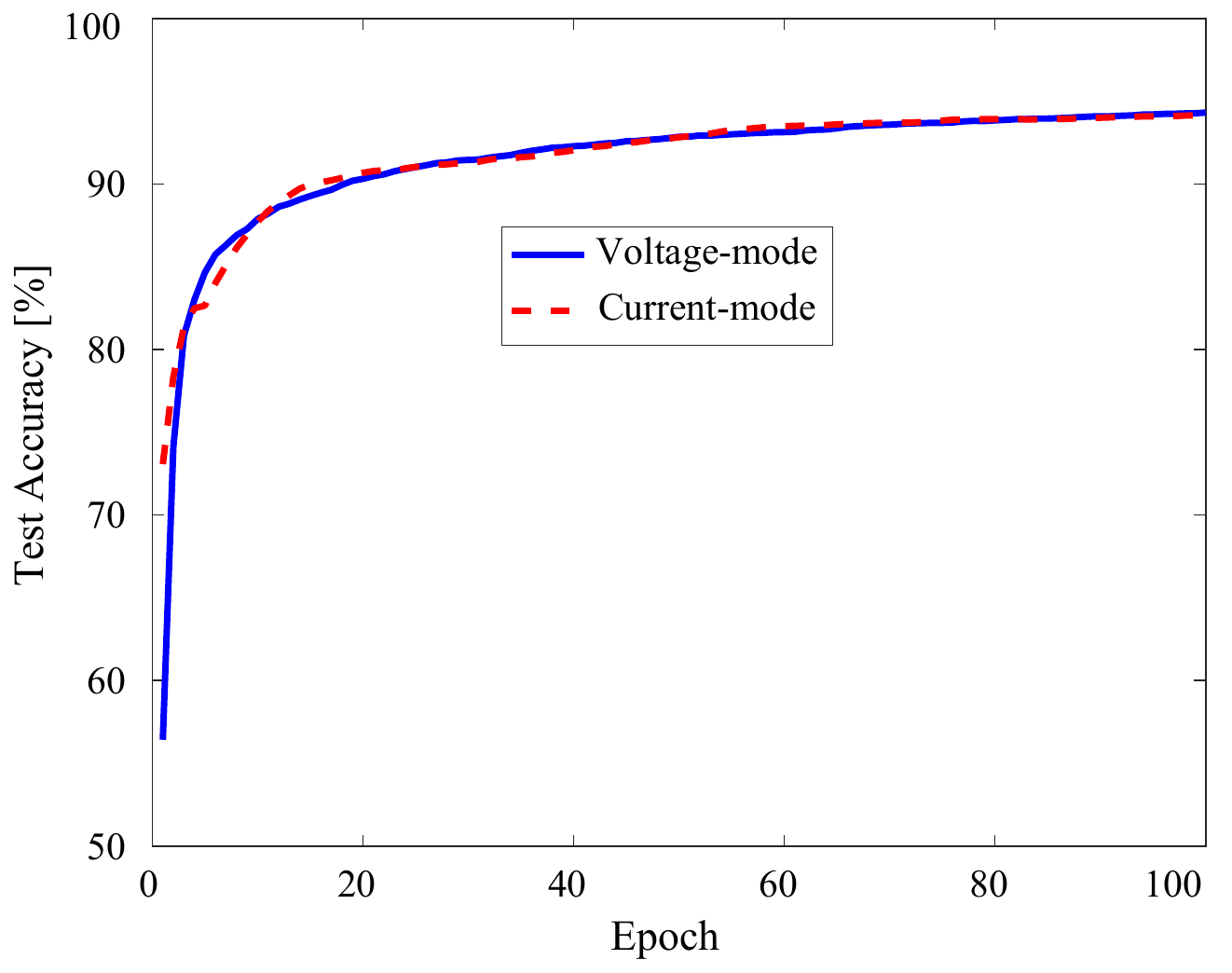}
\caption{Accuracy vs. epoch for the MNIST classification problem.}
\label{fig:mnist_results}
\end{figure}

The MNIST handwritten digit dataset \cite{MNIST}, which consists of 60000 training images and 10000 test images, was used to investigate current-mode crossbars within a NMS.  Each grayscale image was scaled from its original size of 28$\times$28 pixels down to  $7\times7$ pixels using averaging.  Figure \ref{fig:mlp} shows the network architecture, which is a multilayer perceptron (MLP) with a single hidden layer.  $N=49$ is equal to the number of pixels in each MNIST image, while $M=10$ is equal to the number of classes (one for each digit $0,1,\ldots ,9$).  The size of the hidden layer is empirically chosen as $H=50$.  The two weight matrices (input-to-hidden and hidden-to-output weights) are implemented using two memristor crossbars with bipolar weights and dummy rows for the current-mode designs.  The network was trained using online backpropagation.  For the voltage-mode crossbars, this is just the normal backpropagation algorithm.  For current-mode crossbars, the weight updates are given in (\ref{eqn:current_update}).  Figure \ref{fig:mnist_results} shows the classification accuracy on the 10000 test images vs. training epoch.  The voltage-mode, current-mode and simplified current-mode memristor crossbar networks all have similar performance.

\begin{figure}[!t]
\centering
\includegraphics[width=0.65\columnwidth]{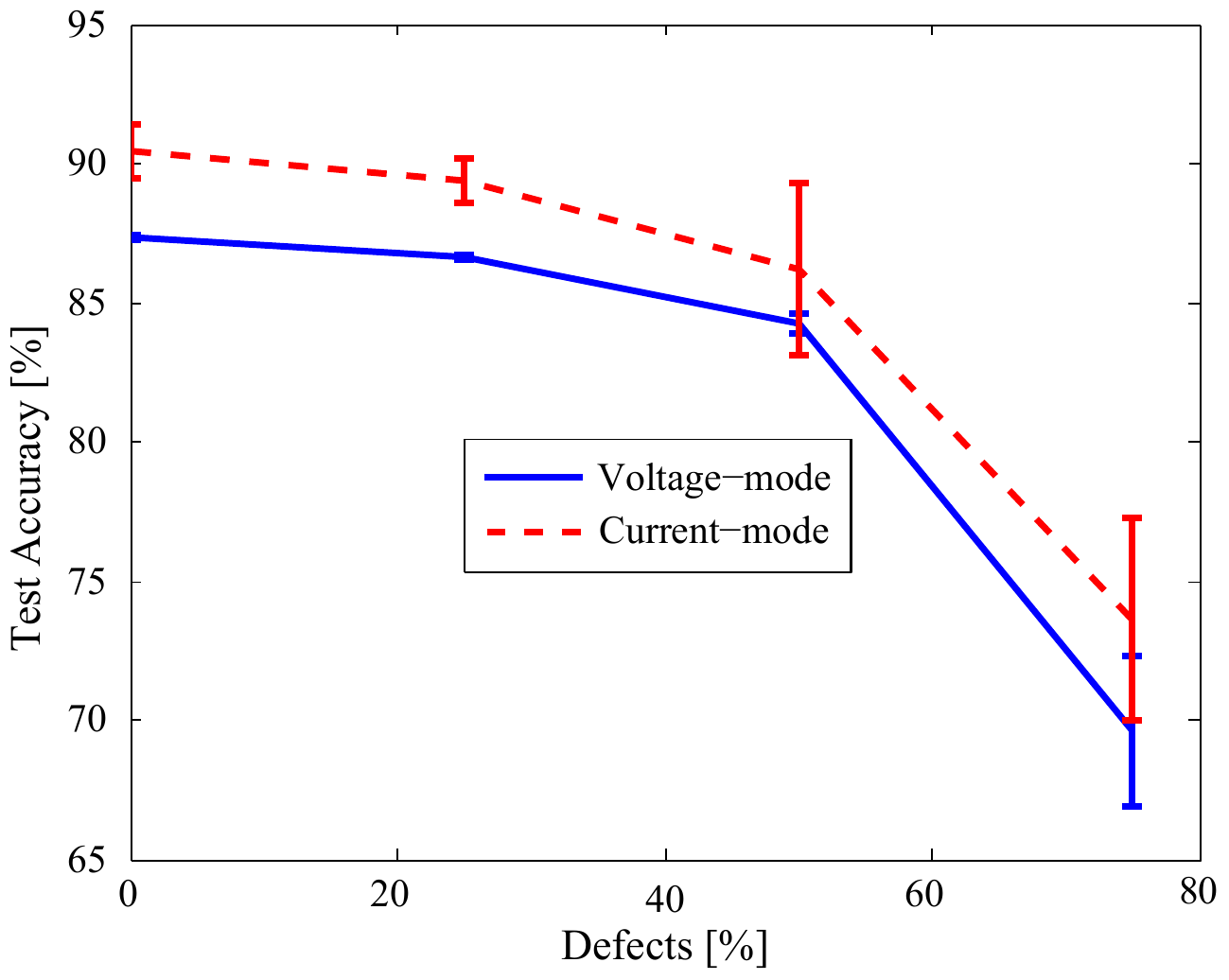}
\caption{Mean accuracy vs. defect rate for the MNIST classification problem.  Error bars indicate standard deviation over 3 runs.}
\label{fig:mnist_defects_results}
\end{figure}

In the voltage-mode crossbar, a defective memristor (i.e. one that has an immutable conductance value) will not affect the weight range corresponding to any other memristors in the crossbar.  However, in the current-mode design, if a device is stuck at a particular conductance, then the weight range of the associated crossbar column will be shifted and scaled.  This difference between the effect of defects in the voltage and current-mode designs motivated a study of the defect tolerance of the two designs.  First, the conductances of the MLP shown in Figure \ref{fig:mlp} were randomly initialized.  Then, a random subset of the memristors was chosen to be \textit{defective}, meaning that their conductance values are immutable.  The MLP was trained on the MNIST dataset with the defective memristors over a range of defect rates from 0\% to 75\%.  The results are shown in Figure \ref{fig:mnist_defects_results}.  Both designs display similar degradation of accuracy as the defect rate increases, indicating that the current-mode design does not have significantly different defect tolerance.

\begin{figure}
\centering
\includegraphics[width=0.63\columnwidth]{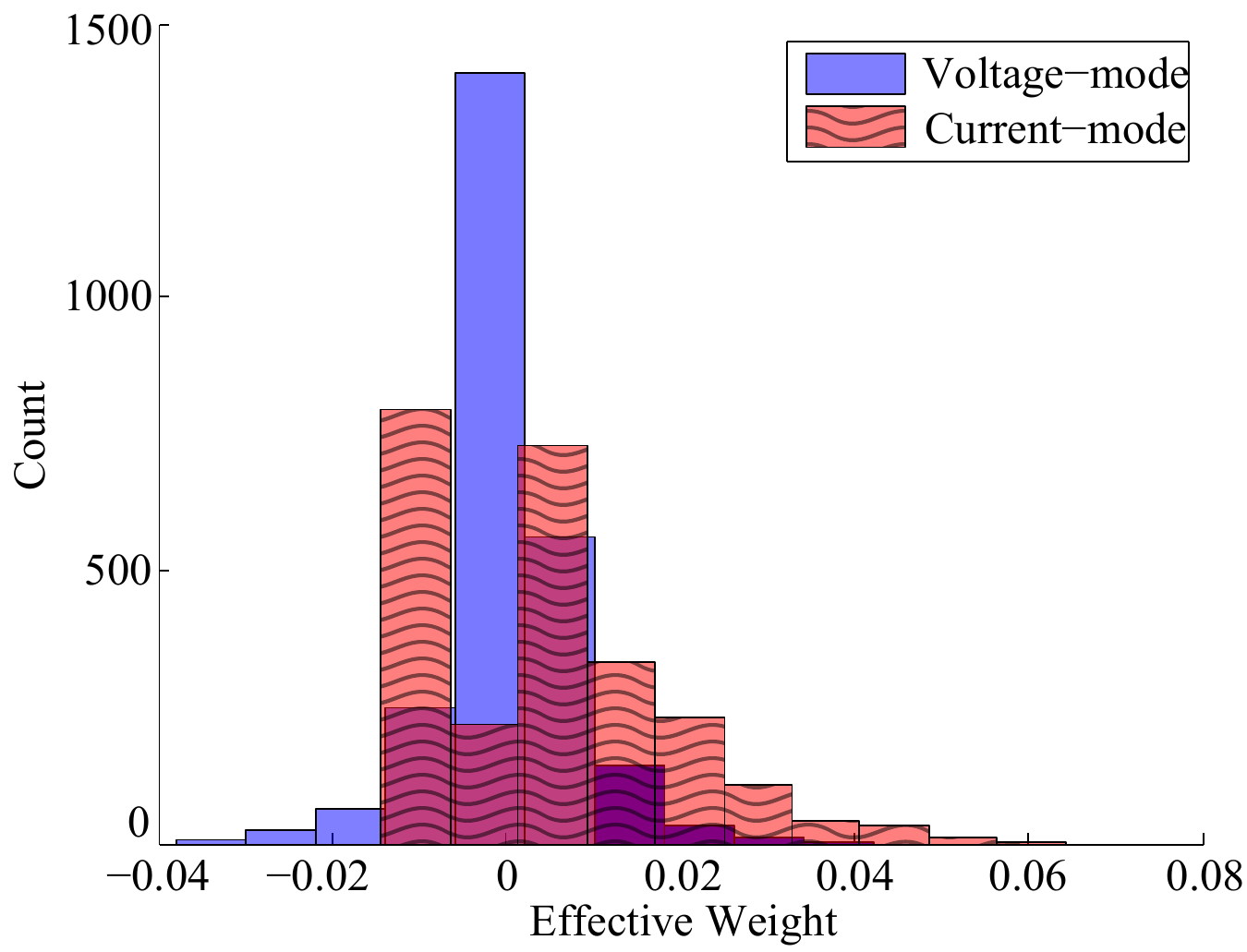}
\caption{Hidden-layer weight distributions for voltage and current-mode NMSs.}
\label{fig:hist}
\end{figure}

Another interesting question is whether the weight constraint in the current-mode crossbar design will yield a different internal representation of features than that of the voltage-mode design.  Figure \ref{fig:hist} shows the weight distributions of the hidden layers (the output layer displayed similar results).  As expected, the voltage-mode weight distributions are Gaussian.  However, the current-mode distributions are asymmetrical with longer tails, indicating that there may be a fundamental difference in the representations, but further study, such as Kullback–Leibler divergence measures or hidden unit statistics, is needed to quantify this.

\section{Conclusions}
\label{section:conclusions}

Current-mode circuits have many attractive properties but, until now, they have not been studied in memristor crossbar-based NMSs.  This paper explored the design of current-mode memristor crossbar circuits to implement weight matrices in NMSs.    Theoretical results were given for comparing the weight range, distribution, and training methods used in NMSs that employ voltage-mode crossbars and current-mode crossbars.  Theoretical and simulation results indicated that NMSs with current-mode crossbars can achieve identical accuracy as voltage-mode designs on learning tasks.  In addition, the defect tolerance of current-mode and voltage-mode crossbar NMSs was shown to be similar, while the final weight distributions of both have obvious qualitative differences.  Future work in this area should focus on the transition from behavioral simulations to circuit-level simulations to study non-ideal effects such as crossbar wire resistance and noise, as well as power consumption.

\section*{Acknowledgments}

The material and results presented in this paper have been CLEARED (Distribution A) for public release, unlimited distribution by AFRL, case number 88ABW-2017-1080. Any opinions, findings and conclusions or recommendations expressed in this material are those of the author and do not necessarily reflect the views of AFRL or its contractors.

\footnotesize
\bibliographystyle{IEEEtran}
\bibliography{library}

\end{document}